\documentclass[runningheads]{llncs}
\usepackage[T1]{fontenc}
\usepackage{booktabs}
\usepackage{multirow}
\usepackage{graphicx}
\usepackage{subcaption}
\usepackage{amsmath}
\usepackage{graphicx,verbatim}
\usepackage{amssymb}

\begin{document}
\title{Experience-Guided Self-Adaptive Cascaded Agents for Breast Cancer Screening and Diagnosis with Reduced Biopsy Referrals}
\titlerunning{Breast Cancer Screening and Diagnosis Agents}
%

\author{Pramit Saha\inst{1} \and 
        Mohammad Alsharid\inst{2}\and 
        Joshua Strong\inst{1} \and 
        J. Alison Noble\inst{1}
}

\institute{Department of Engineering Science, University of Oxford
\and Department of Computer Science, Khalifa University}
%
\authorrunning{Saha et al.}
  
\maketitle              
\begin{abstract}
 We propose an experience-guided cascaded multi-agent framework for \textbf{B}reast \textbf{U}ltrasound \textbf{S}creening and \textbf{D}iagnosis, called \textbf{BUSD-Agent}, that aims to reduce diagnostic escalation and unnecessary biopsy referrals.
Our framework models screening and diagnosis as a two-stage, selective decision-making process. A lightweight \textit{`screening clinic'} agent, restricted to classification models as tools, selectively filters out benign and normal cases from further diagnostic escalation when malignancy risk and uncertainty are estimated as low. Cases that have higher risks are escalated to the \textit{`diagnostic clinic'} agent, which integrates richer perception and radiological description tools to make a secondary decision on biopsy referral. To improve agent performance, past records of pathology-confirmed outcomes along with image embeddings, model predictions, and historical agent actions are stored in a memory bank as structured decision trajectories. For each new case, BUSD-Agent retrieves similar past cases based on image, model response and confidence similarity to condition the agent's current decision policy. This enables retrieval-conditioned in-context adaptation that dynamically adjusts model trust and escalation thresholds from prior experiences without parameter updates. Evaluation across 10 breast ultrasound datasets shows that the proposed experience-guided workflow reduces diagnostic escalation in BUSD-Agent from 84.95\% to 58.72\% and overall biopsy referrals from 59.50\% to 37.08\%, compared to the same architecture without trajectory conditioning, while improving average screening specificity by 68.48\% and diagnostic specificity by 6.33\%.

\keywords{Breast Ultrasound \and In-context Trajectory Learning  \and Experience-conditioned Policy Adaptation \and Cascaded Multi-agent Collaboration}


\end{abstract}
%
%
%
\section{Introduction}

\begin{figure*}[t]
    \centering

    \begin{subfigure}[t]{\textwidth}
        \centering
        \includegraphics[width=1.02\columnwidth]{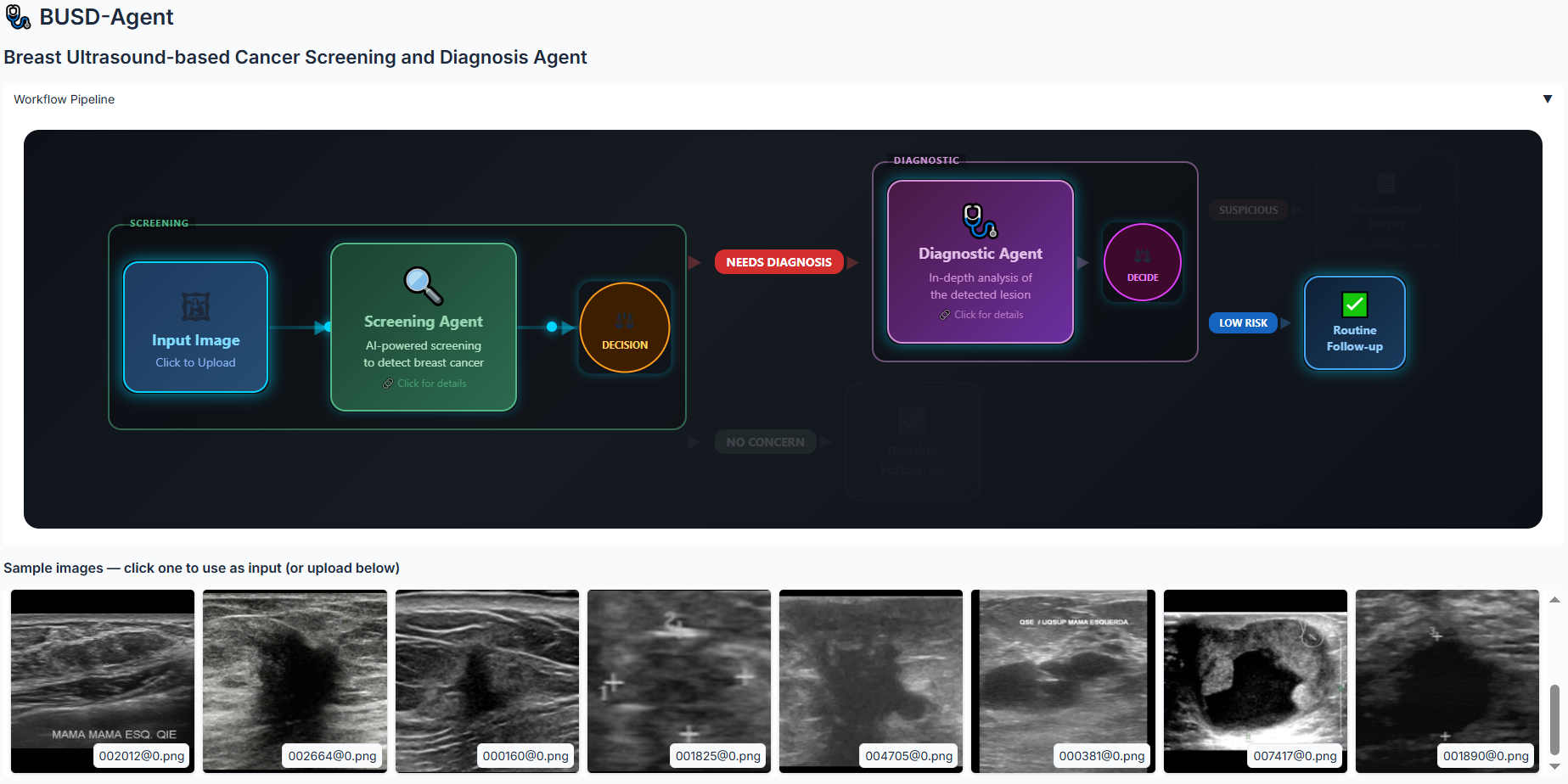}
        \caption{}
        \label{fig:busd_workflow}
    \end{subfigure}

    \begin{subfigure}[t]{0.4\textwidth}
        \centering
        \includegraphics[width=\linewidth]{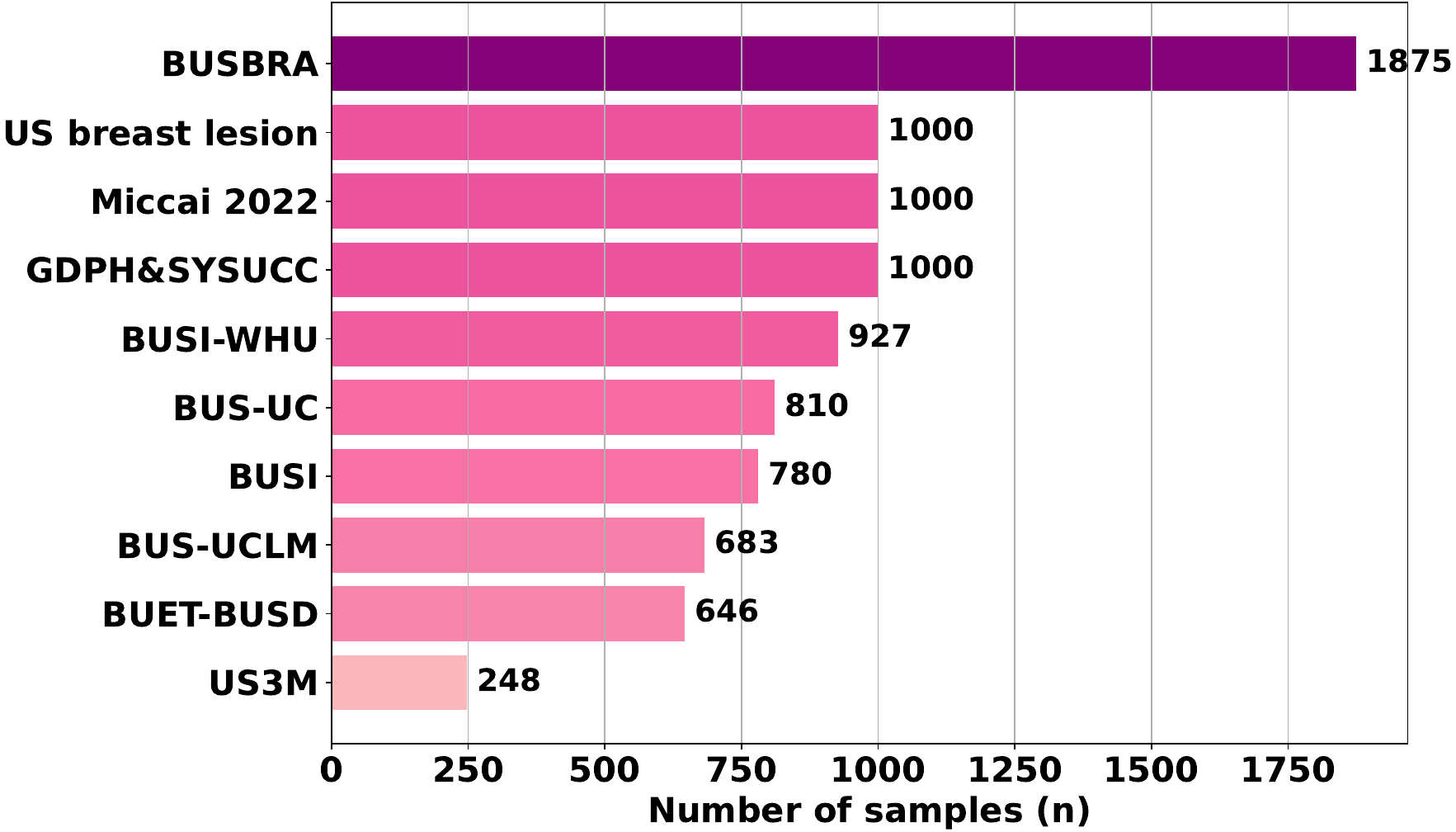}
        \caption{}
        \label{fig:dataset_dist}
    \end{subfigure}\hfill
    \begin{subfigure}[t]{0.6\textwidth}
        \centering
        \includegraphics[width=\linewidth]{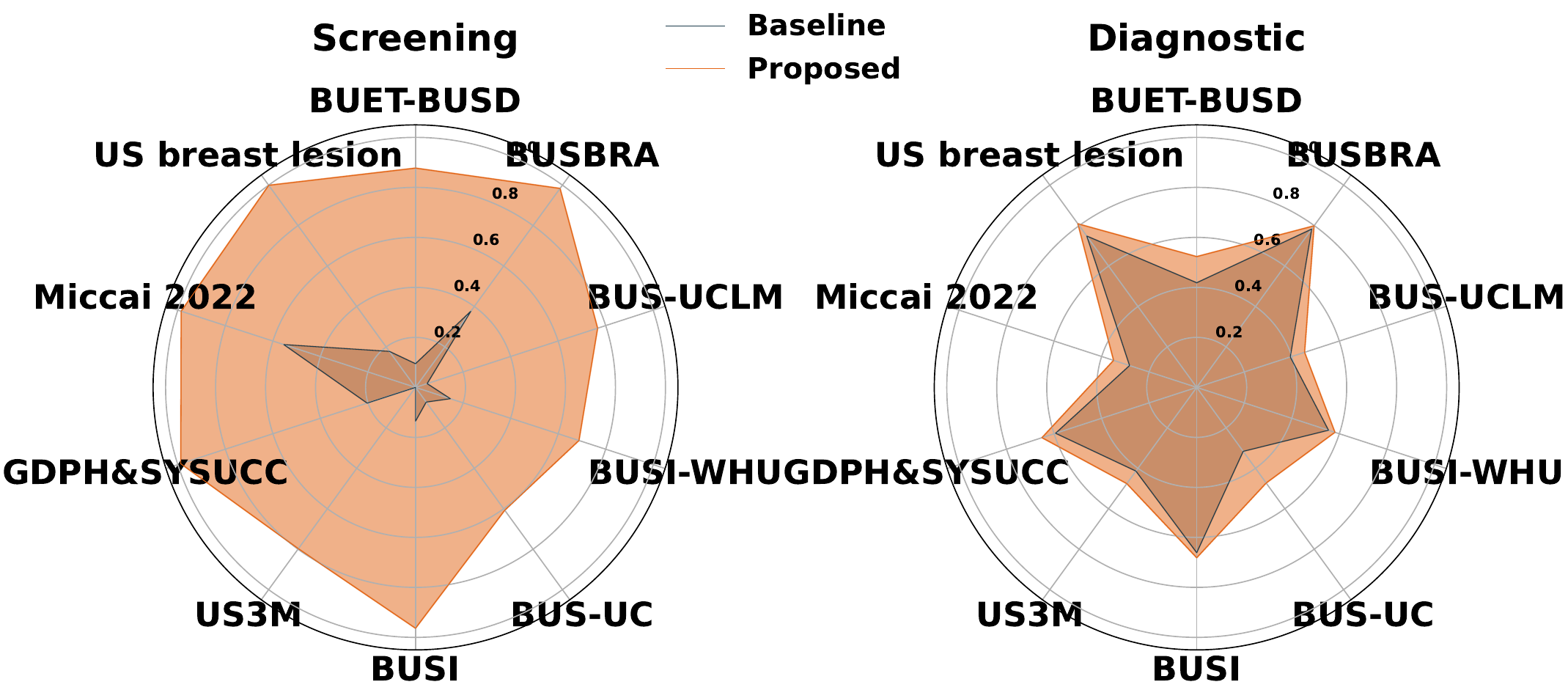}
        \caption{}
        \label{fig:radar_perf}
    \end{subfigure}

    \caption{\textbf{BUSD-Agent overview and results.} (a) BUSD-Agent Real-time User Interface, (b)\# Samples in 10 evaluation datasets, (c) Specificity comparison of baseline vs proposed method for Screening (left) and Diagnostic (right) agents.}
    \label{fig:busd_combined}
\end{figure*}

Breast cancer is the most commonly diagnosed cancer among women worldwide, with approximately 2.3 million new cases and 670,000 deaths reported globally in 2022 \cite{freihat2025global}. Breast ultrasound is widely used as a frontline cancer screening and diagnostic tool, particularly for women with dense breast tissue and in resource-constrained settings \cite{brem2015screening}. However, limited screening specificity frequently results in false-positive findings, excessive diagnostic escalation, and unnecessary biopsy referrals \cite{srivastava2019cancer,springfield1996biopsy,berg2020reducing}. Over the past ten years, referrals to breast diagnostic clinics have increased by nearly 100\% \cite{ellis2024efficient}. Given that the majority of screened cases are ultimately benign or normal, improving specificity without compromising sensitivity is essential to reduce clinical burden, patient anxiety, and healthcare costs as it imposes substantial economic and operational burdens on healthcare systems \cite{world2020screening,jahn2019budget}. Biopsies are invasive procedures associated with direct costs and potential complications, as well as indirect costs such as psychological distress \cite{springfield1996biopsy,wardle1992psychological,fraser1992cost}. Excessive escalation of low-risk cases to diagnostic clinics increases radiologist workload, prolongs reporting times, and consumes limited specialist resources \cite{dembrower2020effect,dembrower2023artificial}. As a result, patients with genuinely suspicious findings face delays in evaluation and treatment initiation. In high-volume screening environments, such workflow congestion compromises timely care for critical patients. Although prior work has traditionally focused on developing ultrasound-based classification and segmentation models \cite{tasnim2024cam,gomez2024bus,vallez2025bus,ye2025dflnet,iqbal2024memory,al2020dataset,yan2024tdf,mo2023hover,lin2022new,pawlowska2024curated}, the aforementioned challenges particularly highlight the need for intelligent, safety-aware triage mechanisms that selectively escalate high-risk cases while confidently clearing low-risk cases.

To address this need, we introduce a novel experience-guided multi-agent framework called Breast Ultrasound Screening and Diagnostic Agent (BUSD-Agent). Our proposed cascaded system consists of a lightweight screening agent that selectively escalates cases to a more powerful diagnostic agent. Unlike conventional systems that rely on fixed decision logic, BUSD-Agent accumulates pathology-confirmed outcomes as structured decision trajectories and conditions its screening and biopsy referral policies on historically relevant prior experiences. By integrating stage-wise selective processing with experience-conditioned policy adaptation, the proposed framework is shown to reduce unnecessary diagnostic escalation and biopsy referrals while preserving sensitivity (see Fig. 1). This design enables self-improving agentic behaviour without parameter updates.

\section{BUSD-Agent} 

\subsection{A Novel Cascaded Multi-Agent Framework}
Breast ultrasound assessment in clinical practice follows a staged workflow, where initial screening determines whether further diagnostic evaluation is required. Inspired by this, we design a cascaded multi-agent framework in the form of (i) a \textit{screening agent} that evaluates all cases, and only those deemed suspicious are escalated to (ii) a \textit{diagnostic agent} for more detailed analysis (see Fig. 2).

\begin{figure*}[t]
    \centering
\includegraphics[width=1.02\columnwidth]{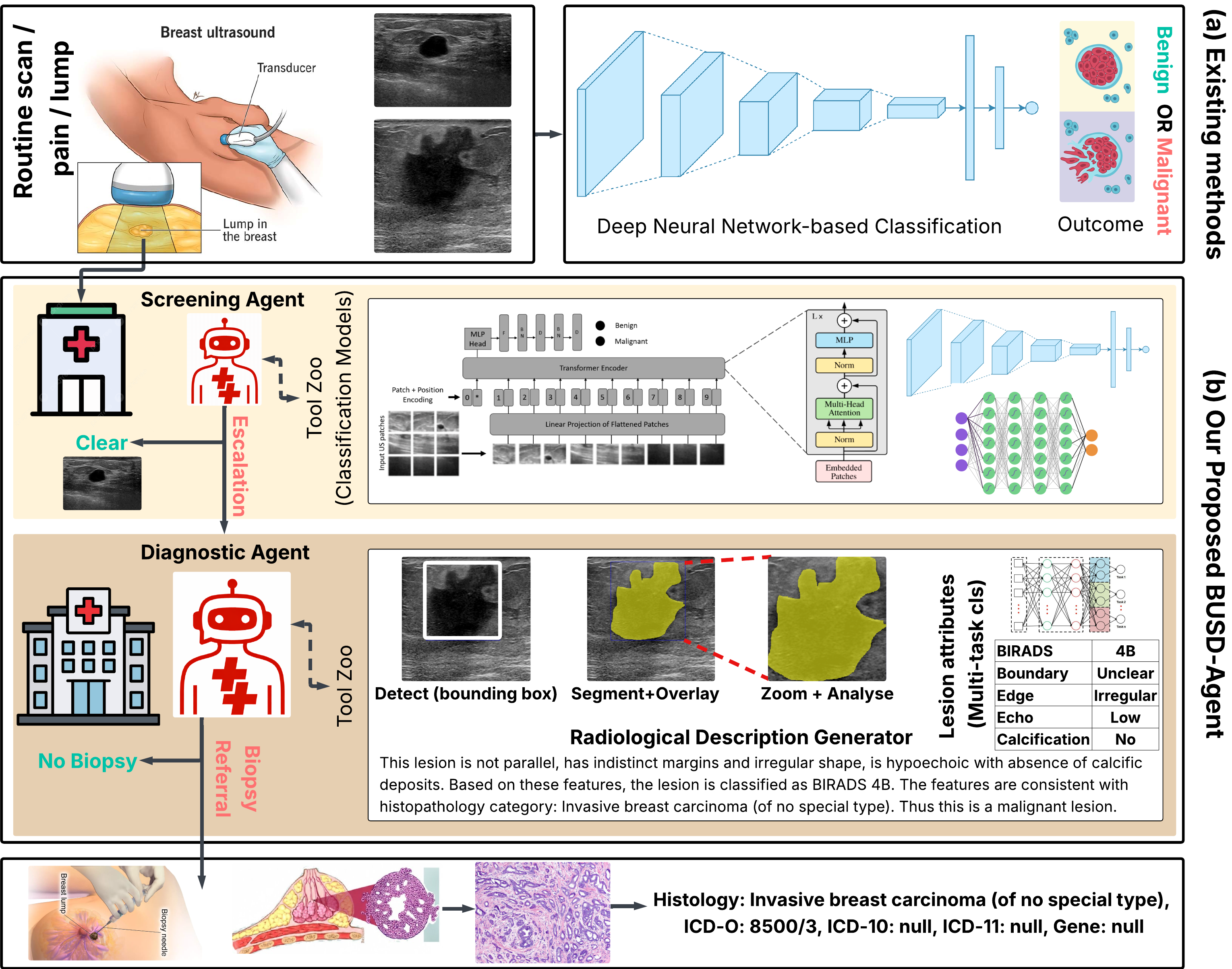}
    \caption{Comparison of our BUSD-Agent Framework with existing methods}
\label{fig:11}
\end{figure*}
\subsubsection{Screening Clinic Agent}
The screening agent operates using a panel of ultrasound-based breast cancer classification models. Given an ultrasound image $x$, the screening agent invokes $M$ independently trained classifiers. Each model $m$ produces a class probability distribution over $\mathcal{C}=\{\texttt{benign}, \texttt{malignant}, \texttt{normal}\}$, from which the predicted class $\hat{y}_m = \arg\max_{c \in \mathcal{C}} p_m(c \mid x)$ is obtained together with its associated confidence score. To provide a global summary of model agreement, an ensemble tool aggregates individual predictions via majority voting: $\hat{y}_{\text{ens}} = \arg\max_{c \in \mathcal{C}} \sum_{m=1}^{M} \mathbb{I}(\hat{y}_m = c)$ where $\mathbb{I}(\cdot)$ denotes the indicator function. The screening toolset returns model predictions, confidence scores, and the ensemble prediction. These structured outputs constitute the complete observation panel available to the screening decision module. The screening agent uses an LLM-based orchestration layer that reconciles these outputs and performs ReAct-style \cite{yao2022react} multi-step reasoning to generate a binary screening decision.

\subsubsection{Diagnostic Clinic Agent}
Cases escalated by the screening agent are processed by a diagnostic clinic agent designed to perform detailed lesion characterization and biopsy decision support. Unlike the screening stage, which relies primarily on classification outputs, the diagnostic agent integrates multiple complementary perception, localization, and description tools. Given an escalated ultrasound image $x$, the diagnostic agent employs an LLM as an orchestration layer that has access to a set of radiological analysis modules, including (i) multi-label predictors for radiological features \textit{viz.}, BIRADS category, lesion edge, boundary, calcification, and echogenicity, (ii) a fine-tuned VLM that produces structured textual reports of lesion morphology, direct spatial analysis tools, including: (iii) bounding-box–based object detection module for lesion localization, (iv) segmentation module that delineates lesion boundaries, (v) zooming and overlay utilities that enable focused inspection of detected or segmented regions. These tools allow the agent to refine spatial attention, examine suspicious regions at higher resolution, and incorporate geometric boundary information into downstream reasoning.  The LLM consumes the integrated diagnostic outputs including radiological feature predictions, localization results, segmentation boundaries, descriptive reports, and screening-stage context and engages a ReAct loop \cite{yao2022react}, iteratively reasoning over these inputs to decompose the task into structured analytical steps that culminate in a final biopsy referral decision.

\begin{figure*}[t]
    \centering
\includegraphics[width=1.02\columnwidth]{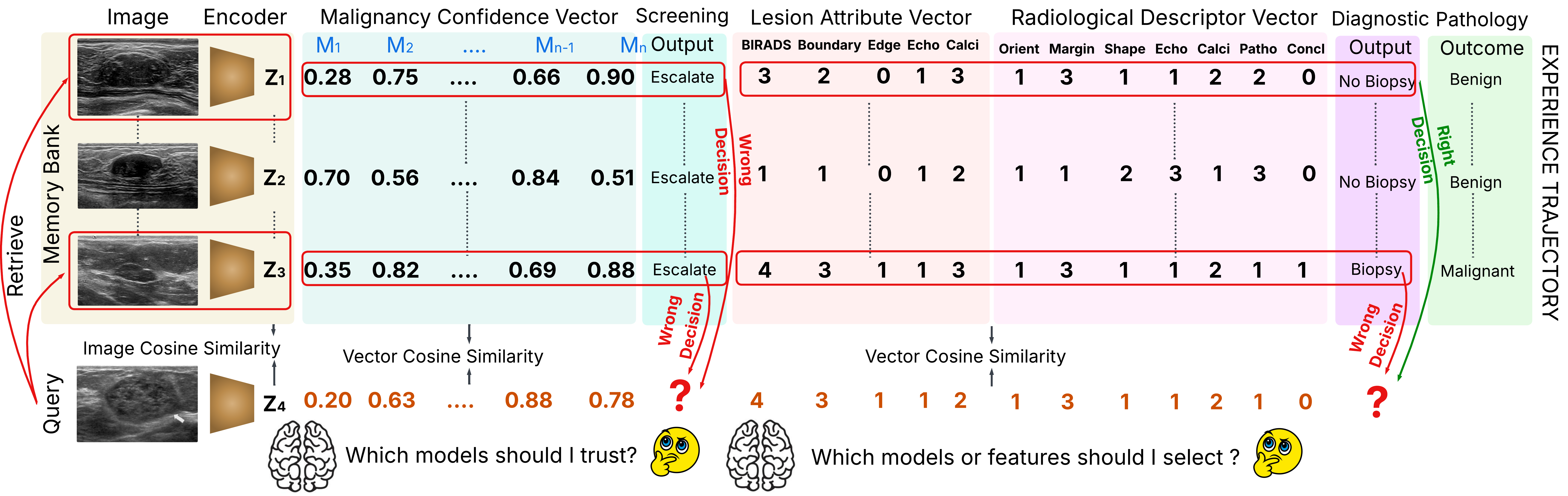}
    \caption{Our experience-guided in-context adaptation strategy for BUSD-Agent}
\label{fig:11}
\end{figure*}
\subsection{Experience-Guided In-Context Adaptation}
Although the baseline framework applies fixed decision logic uniformly across cases, it does not incorporate feedback from prior outcomes to adapt its behaviour. Consequently, the static cascade cannot utilize accumulated successes and failures to recalibrate thresholds, adjust model trust, or refine escalation policies. To address this, we extend the framework with an experience-guided in-context adaptation mechanism grounded in pathology-confirmed prior cases.

\subsubsection{Biopsy-Grounded Memory Bank}
Previous cases with confirmed pathology outcomes are stored as structured decision trajectories in the memory bank. As shown in Fig. 3, each trajectory contains the image embeddings, model predictions, agent decisions, and final biopsy-confirmed labels. This memory bank serves as a repository of historical screening and diagnostic experiences.
\subsubsection{Experience-Conditioned Policy Adaptation for Screening Stage}
For a new case $x$, we retrieve the top-$K$ most similar prior cases from the memory bank using a combined similarity metric defined in a joint feature space based on (a) image embedding $\mathbf{z}(x)$ and (b) a malignancy confidence vector which we define as: $\mathbf{p}(x) = \left(p_1(\texttt{malignant}\mid x), \dots, p_M(\texttt{malignant}\mid x)\right)$ obtained from the $M$ screening classifiers. Similarity between a query case $x$ and a stored case $x_i$ is computed as: $\text{sim}_{\text{screen}}(x, x_i) =
\lambda \cdot \text{cos}(\mathbf{z}(x), \mathbf{z}(x_i))
+ (1-\lambda) \cdot \text{cos}(\mathbf{p}(x), \mathbf{p}(x_i))$ 
where $\text{cos}(\cdot,\cdot)$ denotes cosine similarity and $\lambda \in [0,1]$ is a hyperparameter.  The retrieved set $\mathcal{N}_K(x)$, containing the top-$K$ most similar cases, is formatted as structured in-context exemplar trajectory set and provided as the LLM prior for producing the screening decision. This retrieval mechanism allows the screening agent to examine how visually similar cases with comparable malignancy-confidence profiles were resolved in the past, including whether prior screening decisions aligned with pathology-confirmed outcomes. By conditioning on these exemplar trajectories, the agent can identify which tool outputs or confidence patterns were historically reliable and which led to errors. This allows it to adaptively recalibrate model trust and adjust escalation behaviour for the current case.

\subsubsection{Experience-Conditioned Policy Adaptation for Diagnostic Stage}
For cases escalated to the diagnostic stage, retrieval is performed in a richer feature space reflecting the additional diagnostic outputs. Each diagnostic case is represented by its image embedding $\mathbf{z}(x)$, multi-label radiological feature predictions, and structured categorical outputs derived from the radiological description generator as shown in Fig. 3. Similarity is computed in this joint diagnostic representation to retrieve the top-$K$ most relevant exemplar trajectories. Unlike conventional retrieval approaches that rely solely on visual similarity or free-text descriptors, our diagnostic retrieval operates over structured radiological feature representations, enabling retrieval conditioned on clinically interpretable attributes and decision-relevant cues rather than low-level visual resemblance. The retrieved trajectories capture not only similar radiological patterns but also the downstream consequences of prior biopsy referral decisions. By conditioning on these biopsy-grounded diagnostic experiences, the agent can assess which combinations of radiological features, descriptor patterns, and spatial findings historically led to correct or incorrect biopsy decisions, thereby adaptively refining its biopsy referral policy without parameter updates (see Fig. 3).

\begin{figure*}[t]
    \centering
\includegraphics[width=1.02\columnwidth]{confusion_matrices_screening.png}
    \caption{\textbf{Screening} confusion matrices (\%) for 10 datasets comparing {baseline (w/o experience-conditioning)} and { proposed approach (experience-conditioned)}.}
\label{fig:11}
\end{figure*}

\begin{figure*}[t]
    \centering
\includegraphics[width=1.02\columnwidth]{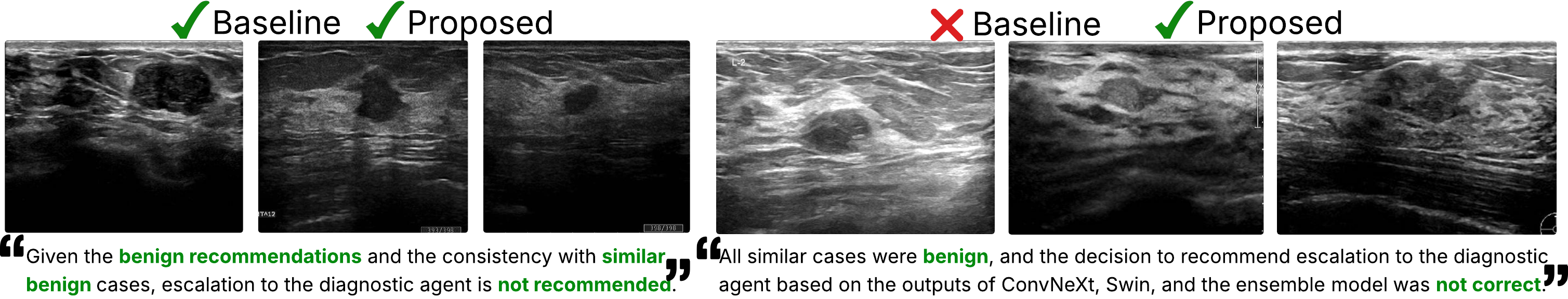}
    \caption{\textbf{Screening Samples}. Left 3 cases were correctly predicted by both. Right 3 cases show harder instances where baseline escalated to diagnostic agent, whereas proposed experience-guided agent correctly identified as benign.}
\label{fig:11}
\end{figure*}

\section{Experiments and Results}
\subsubsection{Datasets and Implementation Details}
We evaluate agentic performance across 10 breast ultrasound datasets: BUET-BUSD \cite{tasnim2024cam}, BUSBRA \cite{gomez2024bus}, BUS-UCLM \cite{vallez2025bus}, BUSI-WHU \cite{ye2025dflnet}, BUS-UC \cite{iqbal2024memory}, BUSI \cite{al2020dataset}, US3M \cite{yan2024tdf}, GDPH\&SYSUCC \cite{mo2023hover}, MICCAI 2022 BUV \cite{lin2022new}, and US Breast Lesion \cite{pawlowska2024curated}.
We use GPT-4o as the orchestration module responsible for tool invocation and structured decision integration. The screening agent consists of 14 classification models spanning CNN- and Transformer-based architectures, including: ConvNeXt (S,T), DeiT (S,T), DenseNet121, EfficientNet-B3, ResNet18, Swin (B,S,T), VGG16, ViT (B,S,T). For the diagnostic stage, the radiological descriptor is implemented using MedGemma-4B fine-tuned with LoRA (r=16) to produce structured ultrasound reports with predefined fields including orientation, margins, shape, echogenicity, calcification, BIRADS, histopathology, and final conclusion. Lesion segmentation and localization are performed using a U-Net with a ResNet50 encoder and YOLOv11 (Ultralytics). In addition, a ResNet50 backbone with 5 task-specific heads is used for lesion attribute recognition: (i) BIRADS (2-5), (ii) edge (Regular/Irregular/Partially Regular), (iii) boundary (Clear/Unclear/Somewhat Unclear/Fairly Clear), (iv) calcification (No/Micro/Suspected/Multiple/Multiple Clustered Micro/Coarse), and (v) echo (Low/Heterogeneous/Cystic-Solid Mixed/ Slightly Low). All models were trained on BUS-CoT dataset \cite{yu2026chain} before integrating into the agentic framework. We select $\lambda = 0.5$, $K = 10$ via hyperparameter sweep and reserve 20\% samples in each dataset to construct the memory bank.

\begin{figure*}[t]
    \centering
\includegraphics[width=1.02\columnwidth]{confusion_matrices_diag.png}
    \caption{\textbf{Diagnostic Agent confusion matrices} (\%) across 10 datasets.}
\label{fig:11}
\end{figure*}

\begin{table*}[t]
\centering
\small
\caption{Benchmarking screening and diagnostic agents across BUS-UCLM and BUS-BRA (BAcc = Balanced Accuracy, Sen = Sensitivity, Spec = Specificity).}
\scalebox{0.88}{
\begin{tabular}{l|ccc|ccc|ccc|ccc}
\hline
\multirow{3}{*}{Model} & \multicolumn{6}{c|}{Screening Clinic Agent} & \multicolumn{6}{c}{Diagnostic Clinic Agent} \\
\cline{2-13}
 & \multicolumn{3}{c|}{BUS-UCLM} & \multicolumn{3}{c|}{BUS-BRA} & \multicolumn{3}{c|}{BUS-UCLM} & \multicolumn{3}{c}{BUS-BRA} \\
\cline{2-13}
 & BAcc & Sen & Spec & BAcc & Sen & Spec & BAcc & Sen & Spec & BAcc & Sen & Spec \\ \hline
\multicolumn{13}{c}{\textbf{Proprietary Models}} \\ \hline
GPT-4o 
& 63.05 & 60.00 & 66.09 
& 57.81 & 27.68 & 87.93 
& 67.62 & 38.46 & 96.77 
& 60.17 & 20.99 & 99.35 \\

GPT-4.1 
& 69.64 & 94.44 & 44.83 
& 68.56 & 78.91 & 58.20 
& 60.20 & 100.00 & 20.41 
& 56.99 & 100.00 & 13.99 \\

GPT-4o-mini 
& 54.89 & 16.67 & 93.10 
& 49.82 & 0.82 & 98.82 
& 50.00 & 100.00 & 0.00 
& 50.00 & 100.00 & 0.00 \\

GPT-4.1-mini 
& 68.31 & 87.78 & 48.85 
& 68.97 & 70.68 & 67.27 
& 48.89 & 0.00 & 97.78 
& 44.59 & 0.00 & 89.19 \\

Gemini 2.5 Pro 
& 62.96 & 100.00 & 25.93 
& 53.17 & 94.87 & 11.48 
& 60.00 & 100.00 & 20.00 
& 68.87 & 100.00 & 37.74 \\

Gemini 2.5 Flash 
& 53.70 & 100.00 & 7.41 
& 50.25 & 92.31 & 8.20 
& 50.00 & 100.00 & 0.00 
& 50.00 & 100.00 & 0.00 \\ \hline

\multicolumn{13}{c}{\textbf{Open-Source Models}} \\ \hline

Qwen 2.5 VL 32B 
& 61.11 & 100.00 & 22.22 
& 52.71 & 92.31 & 13.11 
& 59.52 & 100.00 & 19.05 
& 52.38 & 97.22 & 7.55 \\

Mistral-3.2-24B-Ins
& 53.70 & 100.00 & 7.41 
& 49.68 & 84.62 & 14.75 
& 62.00 & 100.00 & 24.00 
& 80.77 & 100.00 & 61.54 \\

Nemotron-12B-v2-VL 
& 49.16 & 90.91 & 7.41 
& 59.73 & 94.87 & 24.59 
& 52.00 & 100.00 & 4.00 
& 56.52 & 100.00 & 13.04 \\

Llama-3.2-11B-V-Ins 
& 50.00 & 0.00 & 100.00 
& 52.56 & 5.013 & 100.00 
& 42.00 & 0.00 & 84.00 
& 37.50 & 0.00 & 75.00 \\
\hline

\textbf{Ours} 
& 78.29 & 79.75 & 76.62 
& 98.67 & 98.96 & 98.38 
& 71.58 & 97.78 & 45.39 
& 88.96 & 98.02 & 79.90 \\
\hline
\end{tabular}}
\end{table*}

\subsubsection{Performance of Screening Agent}
As illustrated in Fig.~4, evaluation across 10 breast ultrasound datasets shows that the proposed experience-conditioned policy adaptation consistently reduces false positives while maintaining sensitivity relative to the baseline screening agent. On average, a \textbf{25.19\% improvement in true negative rate} is observed, with particularly large gains on BUSBRA (+39.83), BUS-UCLM (+64.13), US Breast Lesion (+31.45), BUSI (+53.75), and US3M (+24.12), reflecting enhanced specificity and more conservative case escalation. Importantly, the method achieves an average \textbf{41.83\% reduction in escalation rate} (\textit{i.e.}, false positives) without substantial degradation in overall true positive rates (-0.15\%), indicating that malignancy detection performance is largely preserved. 
Representative screening examples are shown in Fig.~5, illustrating cases where the proposed mechanism avoids unnecessary escalation compared to the baseline.
\subsubsection{Performance of Diagnostic Agent}
The diagnostic agent consistently \textbf{outperforms} the screening agent in malignancy detection, highlighting the added value of radiological analysis tools. For diagnostic stage, as shown in Fig. 6, the proposed experience-conditioned policy adaptation leads to \textbf{consistent reductions in false positive rates} in BUS-UCLM (-5.26\%), BUS-UC (-6.90\%), Miccai 2022 (-9.98\%), and US breast lesion (-31.16\%), accompanied by corresponding increases in true negative proportions. Importantly, these gains in specificity do not come at the expense of sensitivity: true positive rates remain stable or improve slightly in BUSBRA (+1.79\%), BUSI (+1.43\%). Overall, the experience-conditioning in diagnostic agent demonstrates tighter control of false positives while maintaining acceptable true positive rates which leads to fewer unnecessary biopsy referrals.
\subsubsection{Comparison with SOTA VLMs}
We benchmark BUSD-Agent against proprietary and open-source VLMs (Tab.~1). Across both stages, our framework achieves the \textbf{highest} balanced accuracy with better sensitivity–specificity trade-offs. In contrast, VLMs exhibit unstable operating points, often favouring either sensitivity or specificity. This highlights the benefit of tool-assisted reasoning.

\begin{figure*}[htbp]
    \centering
\includegraphics[width=1.02\columnwidth]{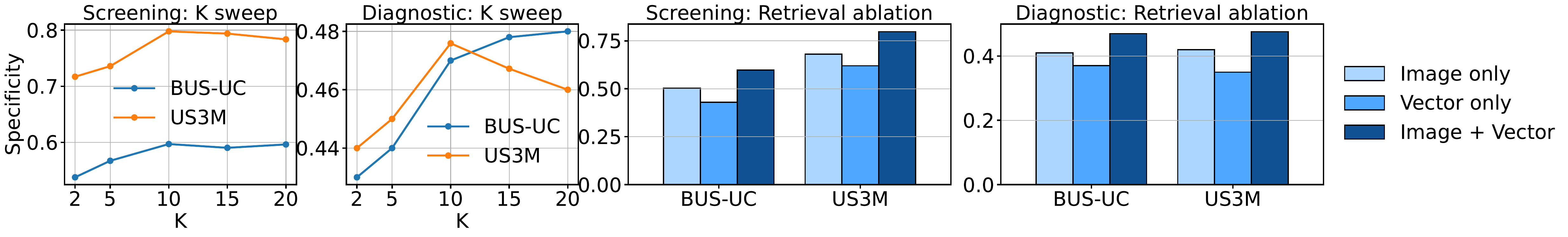}
    \caption{{Effect of \# retrieved exemplars (left) and retrieval strategy (right)}}
\label{fig:11}
\end{figure*}
\subsubsection{Ablation Studies}
 We vary the number of retrieved samples $K \in \{2, 5, 10, 15, 20\}$ in Fig.~7 (left) and observe that performance improves steadily from $K=2$ to $K=10$, after which the curve plateaus or exhibits a small drop. This suggests that incorporating a moderate number of relevant exemplar trajectories is sufficient for stable decision adaptation, with $K=10$ providing the best trade-off between contextual guidance and noise from less similar cases.

Additionally, we show the effectiveness of combined retrieval mechanism by comparing it with image-only and vector–only retrieval in Fig.~7 (right). The combined similarity metric consistently yields higher specificity, indicating that jointly leveraging visual representations and malignancy-confidence profiles results in more informative exemplar selection than either signal.

\section{Discussion and Conclusion}
The primary contributions of our paper are fourfold:
\begin{enumerate}
\item We introduce BUSD-Agent, a cascaded multi-agent framework that: (i) mimics clinical triage by escalating suspicious breast ultrasound cases from screening to diagnostic agent, and (ii) emulates biopsy referral decision by directing potentially malignant cases from diagnostic evaluation to pathology confirmation. We train and develop a suite of underlying tools including malignancy classification, lesion detection, segmentation, characterization, and radiological description generation models that enable the framework. 
\item We address the issue of excessive false positives in screening and diagnostic stages by proposing an experience-conditioned policy adaptation mechanism. Our approach stores past pathology-confirmed cases, including agent actions, model predictions and confidence scores as structured decision trajectories, and performs stage-specific retrieval based on a joint feature space involving images and tool outputs. This conditioning based on retrieved samples enables the agent to make more informed decisions by adaptively adjusting individual model trust and escalation behaviour using prior experiences.

\item We conduct the first benchmarking of 10 proprietary and open-source VLMs highlighting their limitations in screening and diagnostic decision stages.

\item We demonstrate that conditioning agent-based decision policies on past decision trajectories enables adaptive screening and biopsy referral without parameter updates. Empirically, this yields a 25.19\% average improvement in true negative rates and a 41.83\% reduction in escalation rate at the screening stage. At the diagnostic stage, it further reduces false positives across all datasets without collapsing true positive rates, resulting in more balanced sensitivity–specificity trade-offs and fewer unnecessary biopsy referrals.
\end{enumerate}

%
%
%
\bibliographystyle{splncs04}
\bibliography{ref}

\end{document}